%% file: holodiff.tex
\documentclass[10pt,twocolumn,letterpaper]{article}
\usepackage{cvpr}
\usepackage[accsupp]{axessibility}
\usepackage[symbol,hang,flushmargin]{footmisc}
\usepackage{amsmath}
\usepackage{amssymb}
\usepackage{booktabs}
\usepackage{graphicx}
\usepackage{multirow}
\usepackage{todonotes}
\usepackage{xspace}
\usepackage{pifont}
\usepackage{cuted}

\usepackage[pagebackref,breaklinks,colorlinks]{hyperref}
\usepackage{cleveref}
\crefname{section}{Sec.}{Secs.}
\Crefname{section}{Section}{Sections}
\Crefname{table}{Table}{Tables}
\crefname{table}{Tab.}{Tabs.}

\renewcommand{\thefootnote}{\fnsymbol{footnote}}
\newcommand{\name}{\textsc{HoloDiffusion}\xspace}

\def\cvprPaperID{XXXX}
\def\confName{####}
\def\confYear{####}

\newcommand\blfootnote[1]{%
  \begingroup
  \renewcommand\thefootnote{}\footnote{#1}%
  \addtocounter{footnote}{-1}%
  \endgroup
}

\makeatletter
\renewcommand{\paragraph}{%
  \@startsection{paragraph}{4}%
  {\z@}{0.25em}{-1em}%
  {\normalfont\normalsize\bfseries}}
\makeatother

\input{macros}

\title{%
\name: Training a 3D Diffusion Model using 2D Images}
\author{%
Animesh Karnewar\new{$^\star$$^\dagger$}\\
UCL\\
{\tt\small a.karnewar@ucl.ac.uk}
\and
Andrea Vedaldi\\
Meta AI\\
{\tt\small vedaldi@meta.com}
\and
David Novotny$^\star$\\
Meta AI\\
{\tt\small dnovotny@meta.com}
\and
Niloy J. Mitra\\
UCL\\
{\tt\small n.mitra@ucl.ac.uk}
}

\begin{document}
\twocolumn[{%
\renewcommand\twocolumn[1][]{#1}%
\maketitle
\thispagestyle{empty}
\begin{center}
\centering
\vspace{-0.7cm}\url{https://holodiffusion.github.io}
\vspace{0.7cm}
\captionsetup{type=figure}
\includegraphics[width=1.01\linewidth]{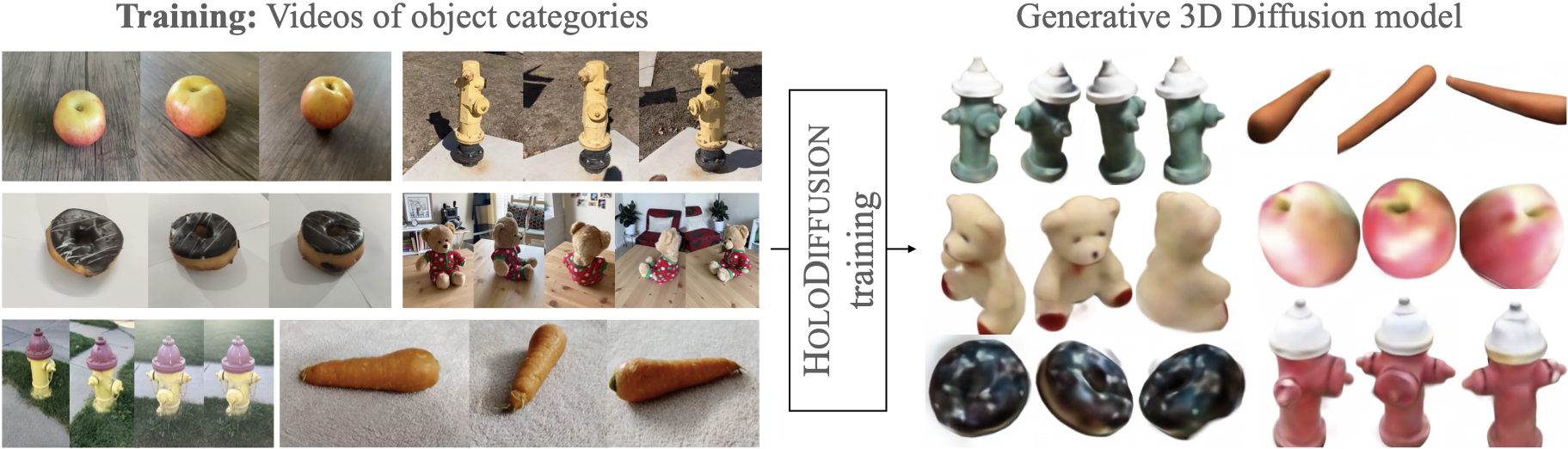}
\captionof{figure}{We present \name as the first \textbf{3D-aware generative diffusion model} that produces 3D-consistent images and is trained with only posed image supervision. Here we show a few different samples generated from models trained on different classes of the CO3D dataset~\cite{reizenstein21common}.}%
\label{fig:teaser-fig}%
\end{center}%
\vspace{1em}%
}]
\input{00_abstract}
\input{01_introduction}

\input{02_related_work}
\input{03_method}
\input{05_experiments}

\input{06_conclusion}
\input{07_acknowledgements}
{\small\bibliographystyle{ieee_fullname}\bibliography{holodiff,vedaldi_general,co3d_bib}}

\cleardoublepage\newpage\appendix
\input{suppl_body.tex}

\end{document}

%% file: macros.tex
\makeatletter
\newcommand{\oneptsmaller}[1]{%
  \begingroup
  \fontsize{\dimexpr\f@size pt-1pt}{\f@baselineskip}\selectfont
  #1%
  \endgroup
}
\makeatother

\newcommand{\real}{\mathbb{R}}

\newcommand{\bx}{\mathbf{x}}
\newcommand{\bp}{\mathbf{p}}
\newcommand{\bc}{\mathbf{c}}
\newcommand{\br}{\mathbf{r}}

\newcommand{\bu}{\mathbf{u}}

\newcommand{\bru}{\br_\bu}

\newcommand{\cmark}{\checkmark}%
\newcommand{\xmark}{\scalebox{0.85}{\ding{53}}}%

\definecolor{colornew}{HTML}{f8c8dc}
\newcommand{\new}[1]{#1}

%% file: 00_abstract.tex
\begin{abstract}%
Diffusion models have emerged as the best approach for generative modeling of 2D images.
Part of their success is due to the possibility of training them on millions if not billions of images with a stable learning objective.
However, extending these models to 3D remains difficult for two reasons.
First, finding a large quantity of 3D training data is much more complex than for 2D images.
Second, while it is conceptually trivial to extend the models to operate on 3D rather than 2D grids, the associated cubic growth in memory and compute complexity makes this infeasible.
We address the first challenge by introducing a new diffusion setup that can be trained, end-to-end, with only posed 2D images for supervision; and the second challenge by proposing an image formation model that decouples model memory from spatial memory.
We evaluate our method on real-world data, using the CO3D dataset which has not been used to train 3D generative models before.
We show that our diffusion models are scalable, train robustly, and are competitive in terms of sample quality and fidelity to existing approaches for 3D generative modeling. 
\end{abstract}

%% file: 01_introduction.tex
\section{Introduction}

\blfootnote{$^\star$ Indicates equal contribution.\\\new{$^\dagger$ part of this work was done during an internship at MetaAI.}}

Diffusion models have rapidly emerged as formidable generative models for images, replacing others (\eg, VAEs, GANs) for a range of applications, including image colorization~\cite{palette:sigg:22}, image editing~\cite{meng2021sdedit}, and image synthesis~\cite{ho2022cascaded,dhariwal2021diffusion}.
These models explicitly optimize the likelihood of the training samples, can be trained on millions if not billions of images, and have been shown to capture the underlying model distribution better~\cite{dhariwal2021diffusion} than previous alternatives.

A natural next step is to bring diffusion models to 3D data.  
Compared to 2D images, 3D models facilitate direct manipulation of the generated content, result in perfect view consistency across different cameras, and allow object placement using direct handles.
However, learning 3D diffusion models is hindered by the lack of a sufficient volume of 3D data for training.
A further question is the choice of representation for the 3D data itself (\eg, voxels, point clouds, meshes, occupancy grids, etc.).
Researchers have proposed 3D-aware diffusion models for point clouds~\cite{pointCloudDiffusion:21}, volumetric shape data using wavelet features~\cite{waveletDiffusion:22} and novel view synthesis~\cite{watson20223dim}.
They have also proposed to distill a pretrained 2D diffusion model to generate neural radiance fields of 3D objects~\cite{poole2022dreamfusion, lin2022magic3d}.
However, a diffusion-based 3D generator model trained using only 2D image for supervision is not available yet.

In this paper, we contribute \name, the first unconditional 3D diffusion model that can be trained with \textit{only} real posed 2D images.
By posed, we mean different views of the same object with known cameras, for example, obtained by means of structure from motion~\cite{schoenberger2016sfm}.

We make two main technical contributions:
(i) We propose a new 3D model that uses a hybrid explicit-implicit feature grid.
The grid can be rendered to produce images from any desired viewpoint and, since the features are defined in 3D space, the rendered images are consistent across different viewpoints.
Compared to utilizing an explicit density grid, the feature representation allows for a lower resolution grid.
The latter leads to an easier estimation of the probability density due to a smaller number of variables.
Furthermore, the resolution of the grid can be decoupled from the resolution of the rendered images.
(ii)
We design a new diffusion method that can learn a distribution over such 3D feature grids while only using 2D images for supervision.
Specifically, we first generate intermediate 3D-aware features conditioned only on the input posed images. 
Then, following the standard diffusion model learning, we add noise to this intermediate representation and train a denoising 3D UNet to remove the noise.
\new{We} apply the denoising loss as photometric error between the rendered images and the Ground-Truth training images.
The key advantage of this approach is that it enables training of the 3D diffusion model from 2D images, which are abundant, sidestepping the difficult problem of procuring a huge dataset of 3D models for training.

We train and evaluate our method on the Co3Dv2~\cite{reizenstein21common} dataset where \name outperforms existing alternatives both qualitatively and quantitatively.

%% file: 02_related_work.tex
\section{Related Work}
\label{s:realted}

\subsection{Image-conditioned 3D Reconstruction}

\paragraph{Neural and differentiable rendering.}

Neural rendering~\cite{tewari2020state} is a class of algorithms that partially or entirely use neural networks to approximate the light transport equation. 

The 2D versions of neural rendering include variants of pix2pix~\cite{isola2017image}, deferred neural rendering~\cite{thies2019deferred}, and their follow-up works.
The common theme in all these methods is that a post-processor neural network (usually a CNN) maps neural feature images into photorealistic RGB images.

The 3D versions of neural rendering have recently been popularized by NeRF~\cite{mildenhall2020nerf}, which uses a Multi Layer Perceptron (MLP) to model the parameters of the 3D scene (radiance and occupancy) and a physically-based rendering procedure (Emission-Absorption raymarching).
NeRF solves the inverse rendering problem where, given many 2D images of a scene, the aim is to recover its 3D shape and appearance.
The success of NeRF gave rise to many follow-up works \cite{mildenhall2020nerf, barron2021mip, zhang2020nerf++, barron2022mip, verbin2022ref}.
While NeRF uses MLPs to represent the occupancy and radiance of the underlying scene, different representations were explored in~\cite{lombardi2019neural, chen2022tensorf, yu2021plenoxels, sun2022direct, muller2022instant, wang2018pixel2mesh, groueix2018papier, karnewar2022relu}.

\begin{figure*}[t!]
\centering
\includegraphics[width=\textwidth]{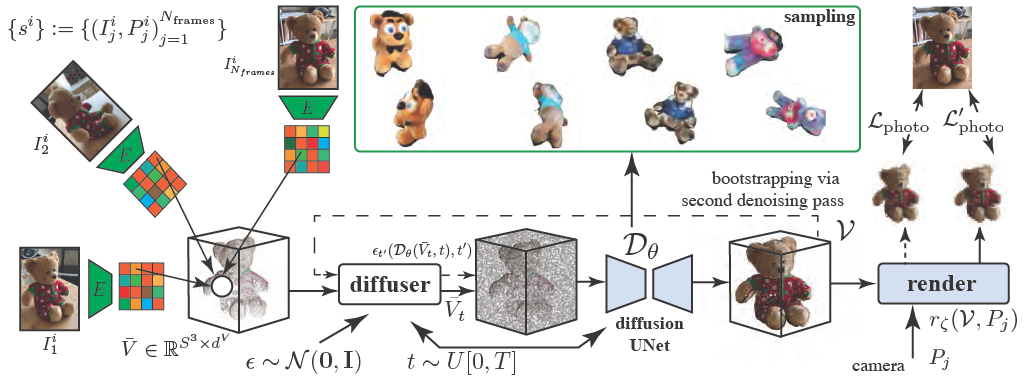}
\caption{
\textbf{Method overview. }
Our \name takes as input video frames for category-specific videos $\{s^i\}$ and trains a diffusion-based generative model $\mathcal{D}_\theta$.
The model is trained with only posed image supervision $\{(I_j^i, P_j^i)\}$, without access to 3D ground-truth.
Once trained, the model can generate view-consistent results from novel camera locations.
Please refer to \cref{sec:method} for details.
}%
\label{fig:pipeline}
\end{figure*}

\paragraph{Few-view reconstruction.}

In many cases, dense image supervision is unavailable, and one has to condition the reconstruction on a small number of scene views instead.
Since 3D reconstruction from few-views is ambiguous, recent methods aid the reconstruction with 3D priors learned by observing many images of an object category. %
Works like CMR~\cite{kanazawa18learning}, C3DM~\cite{novotny20canonical}, and UMR~\cite{li2020self} learn to predict the parameters of a mesh by observing images of individual examples of the object category.
DOVE~\cite{wu21dove:} also predicts meshes, but additionally leverages stronger constraints provided by videos of the deformable objects.

Others~\cite{yu20pixelnerf:,henzler2021unsupervised} aimed at learning to fit NeRF given only a small number of views;
they do so by sampling image features at the 2D projections of the 3D ray samples.
Later works~\cite{reizenstein21common,wang2021ibrnet} have improved this formulation by using transformers to process the sampled features.
Finally, ViewFormer~\cite{kulhanek2022viewformer} drops the rendering model and learns a fully implicit transformer-based new-view synthesizer.
Recently, BANMo~\cite{yang2022banmo} reconstructed deformable objects with a signed distance function.

\subsection{3D Generative Models}

\paragraph{3D Generative advesarial networks.}
Early 3D generative models leveraged adversarial learning~\cite{goodfellow2020gan} as the primary form of supervision.
PlatonicGAN~\cite{henzler2019platonic-gan} learns to generate colored 3D shapes \new{from} an unstructured corpus of images belonging to the same category, by rendering a voxel grid from random viewpoints so that an adversarial discriminator cannot distinguish between the renders and natural images sampled from a large uncurated database.
PrGAN~\cite{gadelha163d-shape} differs from PlatonicGAN by focusing only on rendering untextured 3D shapes.
To deal with the large memory footprint of voxels, HoloGAN~\cite{nguyen-phuoc19hologan:} adjusts PlatonicGAN by rendering a low-resolution 2D feature image of a coarse voxel grid, followed by a 2D convolutional decoder mapping the feature render to the final RGB image.
The results are, however, not consistent with camera motion. 

Inspired by the success of NeRF~\cite{mildenhall20nerf:}, GRAF~\cite{schwarz20graf:} also trains in a data setting similar to PlatonicGAN~\cite{henzler2019platonic-gan} but, differently from PlatonicGAN, represents each generated scene with a neural radiance field.
The GRAF pipeline was subsequently improved by PiGAN~\cite{chan2021pi}, leveraging a SIREN-based~\cite{sitzmann20implicit} architecture.
Similar to HoloGAN, StyleNerf~\cite{gu2021stylenerf} first renders a radiance feature field followed by a convolutional super-resolution network.
EG3D~\cite{chan2022efficient} further improves the pipeline by initially decoding a randomly sampled latent vector to a tri-plane representation followed by a NeRF-style rendering of a radiance field supported by the tri-plane.
EpiGRAF~\cite{skorokhodov2022epigraf} further improves upon the triplane-based 3D generation.
GAUDI~\cite{bautista2022gaudi} also uses the tri-plane while building upon DeVries et. al.~\cite{devries2021unconstrained} which used a single plane representing the floor map of the indoor room scenes being generated.

Besides radiance fields, other shape representations have also been explored.
While VoxGRAF~\cite{schwarz2022voxgraf} replaces the radiance field of GRAF with a sparse voxel grid, StyleSDF~\cite{or2022stylesdf} employs signed distance fields, and Neural Volumes~\cite{lombardi2019neural} propose a novel trilinearly-warped voxel grid.
Wu et al.~\cite{wu2020unsupervised} differentiably render meshes and aid the adversarial learning with a set of constraints exploiting symmetry properties of the reconstructed categories.
Recently, GET3D~\cite{gao2022get3d} differentiably converts an initial tri-plane representation to colored mesh, which is finally rendered.

The aforementioned approaches are trained solely by observing uncurated category-centric image collections \textit{without} the need for any explicit 3D supervision in form of the ground truth 3D shape or camera pose.
However, since the rendering function is non-smooth under camera motion, these methods can either successfully reconstruct image databases with a very small variation in camera poses (\eg, fronto-parallel scenes such as portrait photos of cat or human faces) or datasets with a well-defined distribution of camera intrinsics end extrinsics (\eg, synthetic datasets).
We tackle the pose estimation problem by leveraging a dataset of category-centric videos each containing multiple views of the same object.
Observing each object from a moving vantage point allows for estimating accurate scene-consistent camera poses that provide strong constraints.

While most approaches focus on generating shapes of isolated instances of object categories, GIRAFFE~\cite{niemeyer21giraffe:} and BlockGAN~\cite{nguyen-phuoc20blockgan:} extend GRAF and HoloGAN to reconstruct compositions of objects and their background.
Alternative approaches focus on text-conditioned 3D shape generation~\cite{bautista2022gaudi,jain2022zero,poole2022dreamfusion}, or learn~\cite{karnewar_3InGan_3dv_22} a generative model by observing a single self-similar scene.

\paragraph{3D diffusion models.}

Diffusion models for 3D shape learning have been explored only very recently.
Luo~\etal~\cite{luo2021diffusion} use full 3D supervision to learn a generative diffusion model of point clouds.
In a concurrent effort, Watson~\etal~\cite{watson2022novel} learns a new-view synthesis function which, conditioned on a posed image of a scene, generates a new view of the scene from a specified target viewpoint.
Differently from us, \cite{watson2022novel} do not employ an explicit image formation model which may lead to geometrical inconsistencies between generated viewpoints.

%% file: 03_method.tex
\section{\name}%
\label{sec:method}

We start by discussing the necessary background and notation on diffusion models in \cref{s:background}, and then we introduce our method in \cref{s:3ddiff}, \cref{s:BLDM}, and \cref{s:details}.

\subsection{Diffusion Models}\label{s:background}

Given $N$ i.i.d.~samples $\{x^i\}_{i=1}^{N}$ from (an unknown) data distribution $p(x)$, the task of generative modeling is to find the parameters $\theta$ of a parametric model $p_\theta(x)$ that best approximate $p(x)$.
\emph{Diffusion models} are a class of likelihood-based models centered on the idea of defining a forward diffusion (noising) process $q(x_t | x_{t-1})$, for $t \in [0, T]$.
The noising process converts the data samples into pure noise, \ie, $q(x_T) \approx q(x_T | x_{T-1}) = \mathcal{N}(0, I)$.
The model then learns the reverse process $p(x_{t-1} | x_t)$, which iteratively converts the noise samples back into data samples starting from the purely Gaussian sample $x_T$.

The Denoising Diffusion Probabilistic Model (DDPM)~\cite{ho2020denoising}, in particular, defines the noising transitions using a Gaussian distribution, setting
\begin{equation}
q(x_t | x_{t-1}) := \mathcal{N}(x_t; \sqrt{\alpha_t}x_{t-1}, (1-\alpha_t)I). %
\end{equation}
Samples can be easily drawn from this distribution by using the reparameterization trick:
\begin{equation}\label{eqn:x_t|x_t-1}
x_t := \sqrt{\alpha_t}x_{t-1} + \sqrt{1-\alpha_t}\epsilon
~~~\text{where}~~~
\epsilon \sim \mathcal{N}(0, I).
\end{equation}
One similarly defines the reverse denoising step using a Gaussian distribution:
\begin{equation}\label{eq:denoiser_introduction}
p_{\theta}(x_{t-1} | x_t)
:=
\mathcal{N}(x_{t-1}; \sqrt{\alpha_{t}}\mathcal{D}_{\theta}(x_{t}, t), (1-\alpha_t)I),
\end{equation}
where, the $\mathcal{D}_{\theta}$ is the denoising network with learned parameters $\theta$.
The sequence $\alpha_t$ defines the noise schedule as:
\begin{equation}
\alpha_t = 1 - \beta_t,
\quad \beta_t \in [0, 1],
\quad \text{s.t.~}
\beta_t > \beta_{t-1} \forall t \in [0, T].
\end{equation}
We use a linear time schedule with $T=1000$ steps.

The denoising must be applied iteratively for sampling the target distribution $p(x)$.
However, for training the model we can draw samples $x_t$ directly from $q(x_t | x_0)$ as:
\begin{align}
x_t = \sqrt{\bar\alpha_t}x_0 + \sqrt{1 - \bar\alpha_t}\epsilon \quad
\text{where, } \bar\alpha_t = \prod_{i=0}^{t}\alpha_i.
\end{align}
It is common to use the network $\mathcal{D}_\theta(x_t, t)$ to predict the noise component $\epsilon$ instead of the signal component $x_{t-1}$ in \cref{eqn:x_t|x_t-1}; which has the interpretation of modeling the score of the marginal distribution $q(x_t)$ up to a scaled constant~\cite{ho2020denoising, luo2022understanding}.
Instead, we employ the ``$x_0$-formulation'' from \cref{eq:denoiser_introduction}, which has recently been explored in the context of diffusion model distillation~\cite{salimans2022progressive} and modeling of text-conditioned videos using diffusion \cite{ho2022imagen}.
The reasoning behind this design choice will become apparent later.

\paragraph{Training.}

Training the ``$x_0$-formulation'' of a diffusion model $\mathcal{D}_{\theta}$ comprises minimizing the following loss:
\begin{equation} \label{eq:loss_diffusion_denoise}
\mathcal{L} = \| \mathcal{D}_{\theta}(x_t, t) - x_0 \|^2,
\end{equation}
encouraging $\mathcal{D}_{\theta}$ to denoise sample
$
x_t \sim \mathcal{N}(\sqrt{\bar\alpha_t}x_0, (1 - \bar\alpha_{t})I)
$
to predict the clean sample $x_0$.
 
\paragraph{Sampling.}
Once the denoising network $\mathcal{D}_{\theta}$ is trained, sampling can be done by first starting with pure noise, \ie, $x_T \sim \mathcal{N}(0, I)$, and then iteratively refining $T$ times using the network $\mathcal{D}_{\theta}$, which terminates with a sample from target data distribution $x_0 \sim q(x_0) = p(x)$:
\begin{equation}\label{eq:sampling_diffusion}
x_{t-1} \sim \mathcal{N}(\sqrt{\bar\alpha_{t-1}} \mathcal{D}_{\theta}(x_t, t), (1 - \bar\alpha_{t-1})I).
\end{equation}

\subsection{Learning 3D Categories by Watching Videos}\label{s:3ddiff}

\paragraph{Training data.}

The input to our learning procedure is a dataset of $N \in \mathbb{N}$ video sequences
$
\{ s^i \}_{i=1}^{N}
$,
each depicting an instance of the same object category (\eg, car, carrot, teddy bear).
Each video
$
s^i = (I^i_j, P^i_j)_{j=1}^{N_\text{frames}}
$
comprises $N_\text{frames}$ pairs $(I^i_j, P^i_j)$, each consisting of an RGB image
$
I^i_j \in \real^{3 \times H \times W}
$ 
and its corresponding camera pose $P^i_j \in \real^{4 \times 4}$, represented as a $4 \times 4$ camera matrix.

Our goal is to train a generative model $p(V)$ where $V$ is a representation of the shape and appearance of a 3D object;
furthermore, we aim to learn this distribution using only the 2D training videos $\{ s^i \}_{i=1}^N$.

\paragraph{3D feature grids.}
As 3D representation $V$ we pick \emph{3D feature voxel grids}
$
V \in \real^{d^V \times S\times S\times S}
$
of size $S \in \mathbb{N}$ containing $d^V$-dimensional latent feature vectors.
Given the voxel grid $V$ representing the object from a certain video $s$, we can reconstruct any frame
$
(I_j,P_j) \in s
$
of the video as
$
I_j = r_\zeta(V, P_j)
$
by the means of the \emph{rendering function}
$
r_\zeta(V, P_j): 
\real^{d^V \times S\times S\times S} \times \real^{4 \times 4} 
\mapsto 
\real^{3 \times H \times W}
$,
where $\zeta$ are the function parameters (see \cref{s:details} for details).

Next, we discuss how to build a diffusion model for the distribution $p(V)$ of feature grids.
One might attempt to directly apply the methodology of \cref{s:background}, setting $x=V$, but this does not work because we have no access to ground-truth feature grids $V$ for training; instead, these 3D models must be inferred from the available 2D videos while training.
We solve this problem in the next section.

\subsection{Bootstrapped Latent Diffusion Model}
\label{s:BLDM}

In this section, we show how to learn the distribution $p(V)$ of feature grids from the training videos $s$ alone.
In what follows, we use the symbol $\mathcal{V}$ as a shorthand for $p(V)$.

The training videos provide RGB images $I$ and their corresponding camera poses $P$, but no sample feature grids $V$ from the target distribution $\mathcal{V}$.
As a consequence, we also have no access to the noised samples
$
V_t \sim \mathcal{N}(\sqrt{\bar\alpha_{t}}V_0, (1 - \bar\alpha_{t}) I)
$
required to evaluate the denoising objective \cref{eq:loss_diffusion_denoise} and thus learn a diffusion model.

To solve this issue, we introduce the BLDM (\emph{Bootstrapped Latent Diffusion Model}).
BLDM can learn the denoiser-cum-generator
$
\mathcal{D}_\theta
$ 
given samples $\bar V \sim \mathcal{\bar V}$ from an \emph{auxiliary distribution} $\mathcal{\bar V}$, which is closely related but not identical to the target distribution $\mathcal{V}$.

\paragraph{The auxiliary samples $\bar V$.} %

\new{As} shown in \cref{fig:pipeline}, our idea is to obtain the auxiliary samples $\bar V$ as a (learnable) function of the corresponding training videos $s$.
To this end, we use a design strongly inspired by Warp-Conditioned-Embedding (WCE)~\cite{henzler2021unsupervised}, which demonstrated compelling performance for learning 3D object categories.
Specifically, given a training video $s$ containing frames $I_j$, we generate a grid
$
\bar V \in \real^{d^V \times S \times S \times S}
$
of auxiliary features
$
\bar V_{:mno} \in [-1, 1]^{d_V}
$
by projecting the 3D coordinate $\bx^V_{mno}$ of the each grid element $(m,n,o)$  to every video frame $I_j$, sampling corresponding 2D image features, and aggregating those into a single $d_V$-dimensional descriptor per grid element.
The 2D image features are extracted by a trainable encoder (\new{we use the ResNet-32 encoder}~\cite{he2016deep}) $E$.
This process is detailed in the supplementary material.

\paragraph{Auxiliary denoising diffusion objective.}

The standard denoising diffusion loss \cref{eq:loss_diffusion_denoise} is unavailable in our case because the data samples $V$ are unavailable.
Instead, we leverage the ``$x_0$-formulation'' of diffusion to employ an alternative diffusion objective which does not require knowledge of $V$.
Specifically, we replace \cref{eq:loss_diffusion_denoise} with a \emph{photometric loss}
\begin{equation} \label{eq:loss_aux_diffusion_denoise}
\mathcal{L}_\text{photo} := \| r_\zeta(\mathcal{D}_\theta(\bar V_t, t), P_j) - I_j \|^2,
\end{equation}
which compares the rendering $r_\zeta(\mathcal{D}_\theta(\bar V_t, t), P_j)$
of the denoising
$\mathcal{D}_\theta(\bar V_t, t)$
of the noised auxiliary grid $\bar V_t$ to the (known) image $I_j$ with pose $P_j$.
\Cref{eq:loss_aux_diffusion_denoise} can be computed because the image $I_j$ and camera parameters $P_j$ are known and $\bar V_t$ is derived from the auxiliary sample $\bar V$, whose computation is given in the previous section.

\paragraph{Train/test denoising discrepancy.}

Our denoiser $\mathcal{D}_\theta$ takes as input a sample $\bar V_t$ from the noised \textit{auxiliary} distribution $\bar{\mathcal{V}}_t$ instead of the noised \emph{target} distribution $\mathcal{V}_t$.
While this allows to learn the denoising model by minimizing \cref{eq:loss_aux_diffusion_denoise}, it prevents us from drawing samples from the model at test time.
This is because, during training,  $\mathcal{D}_\theta$ learns to denoise the auxiliary samples $ \bar V \in \mathcal{\bar V}$ (\new{obtained} through fusing image features into a voxel-grid), but at test time we need instead to draw target samples $V \in \mathcal{V}$ as specified by \cref{eq:sampling_diffusion} per sampling step.
We address this problem by using a bootstrapping technique that we describe next.
    
\paragraph{Two-pass diffusion bootstrapping.}

\begin{figure*}
\centering%
\includegraphics[width=0.95\linewidth]{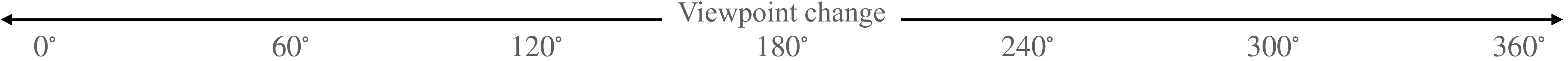}\\\vspace{0.03cm}
\includegraphics[width=0.99\linewidth]{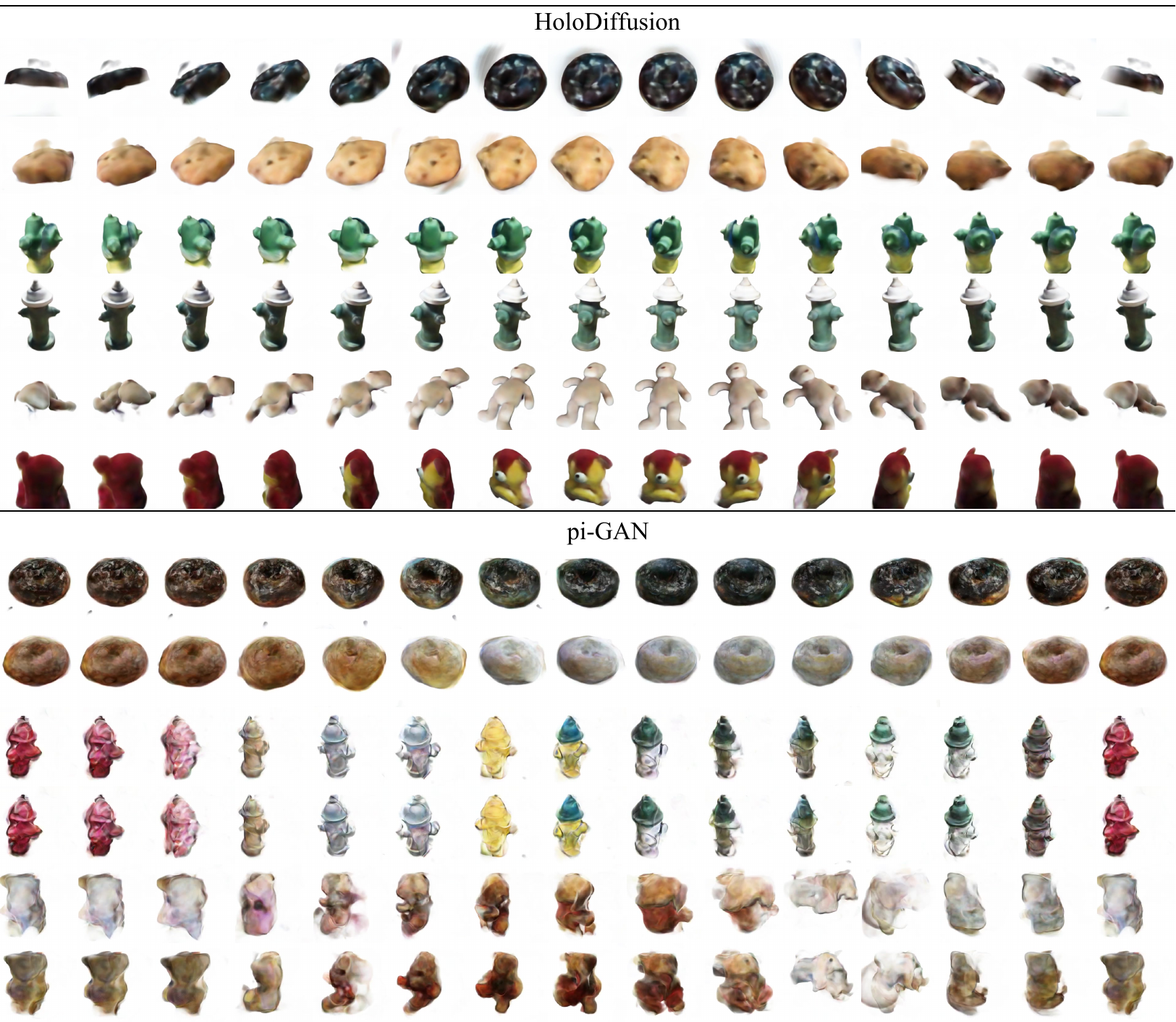}%
\caption{
\textbf{View consistency. }
Evaluation of the consistency of the shape renders under camera motion.
While our results~(top) remain consistent,  pi-GAN~\cite{chan2021pi}'s results~(bottom) suffer from significant appearance variations across view changes.
\label{fig:qualitative_vp}
}
\end{figure*}

In order to remove the discrepancy between the training and testing sample distributions for the denoiser $\mathcal{D}_\theta$, we first use the latter to obtain `clean' voxel grids from the training videos during an initial denoising phase, and then apply a diffusion process to those, finetuning $\mathcal{D}_\theta$ as a result.

Our bootstrapping procedure rests on the assumption that once $\mathcal{L}_\text{photo}$ is minimized,
the denoisings $\mathcal{D}_\theta(\bar V_t, t)$ of the auxiliary grids $\bar V \sim \mathcal{\bar V}$ follow the clean data distribution $\mathcal{V}$, \ie, 
$\mathcal{D}_{\theta^\star}(\bar V_t, t) \sim \mathcal{V}$
for the optimal denoiser parameters $\theta^\star$ that minimize $\mathcal{L}_\text{photo}$. Simply put, the denoiser $\mathcal{D}_\theta$ learns to denoise \textit{both} the diffusion noise and the noise resulting from imperfect reconstructions.
Note that our assumption $\mathcal{D}_{\theta^\star}(\bar V_t, t) \sim \mathcal{V}$ is reasonable since recent single-scene neural rendering methods \cite{mildenhall20nerf:,lombardi2019neural,chen2022tensorf} have demonstrated successful recovery of high-quality 3D shapes solely by optimizing the photometric loss via differentiable rendering.

Given that $\mathcal{D}_{\theta^\star}$ is now capable of generating clean data samples, we can expose it to the noised version of the clean samples $V$ by executing a second denoising pass in a recurrent manner.
To this end, we define the \emph{bootstrapped photometric loss} $\mathcal{L}'_\text{photo}$:
\begin{equation} \label{eq:two_pass_bootstrapp}
\mathcal{L}'_\text{photo} := \| r_\zeta(\mathcal{D}_\theta(\epsilon_{t'}(\mathcal{D}_\theta(\bar V_t, t), t'), P_j)  - I_j \|^2,
\end{equation}
with $\epsilon_{t'}(Z) \sim \mathcal{N}(\sqrt{\bar\alpha_{t'}} Z, (1 - \bar\alpha_{t'})I)$ denoting the diffusion of input grid $Z$ at time $t'$.
Intuitively, \cref{eq:two_pass_bootstrapp} evaluates the photometric error between the ground truth image $I$ and the rendering of the doubly-denoised grid $
\mathcal{D}_\theta(\epsilon_{t'}(\mathcal{D}_\theta(\bar V_t, t), t'))
$.

\subsection{Implementation Details}%
\label{s:details}

\paragraph{Training details.}

\name training finds the optimal model parameters $\theta, \zeta$ by minimizing the sum of the photometric and the bootstrapped photometric losses
$
\mathcal{L}_\text{photo} + \mathcal{L}'_\text{photo}
$
using the Adam optimizer with an initial learning rate 
$5\cdot 10^{-5}$ (decaying ten-fold whenever the total loss plateaus) until convergence is reached.

In each training iteration, we randomly sample 10 source views $\{ I_j \}$ from a randomly selected training video $s^i$ to form the grid of auxiliary features $\bar V$. 
The auxiliary features are noised to form $\bar V_t$ and later denoised with $\mathcal{D}_\theta(\bar V_t)$.
Afterwards $\mathcal{D}_\theta(\bar V_t)$ is noised and denoised again during the two-pass bootstrap procedure.
To avoid two rendering passes in each training iteration (one for $\mathcal{L}_\text{photo}$ and the second for $\mathcal{L}'_\text{photo}$), we randomly choose to optimize $\mathcal{L}'_\text{photo}$ with 50-50 probability in each iteration as a lazy regularization.
The photometric losses compare renders $r_\zeta(\cdot, P_j)$ of the denoised voxel grid to 3 different target views (different from the source views).

\paragraph{Rendering function $r_\zeta$.}

The differentiable rendering function $r_\zeta(V, P_j)$ from \cref{eq:loss_aux_diffusion_denoise,eq:two_pass_bootstrapp} uses Emission-Absorption (EA) ray marching as follows.
First, given the knowledge of the camera parameters $P_j$, a ray $\bru \in \mathcal{S}^2$ is emitted from each pixel $\bu \in \{0, \dots, H-1\} \times \{0, \dots, W-1\}$ of the rendered image $\hat I_j \in \real^{3 \times H \times W}$. 
We sample 
$N_S$ 3D points \new{$(\bp_i)_{i=1}^{N_S}$} on each ray at regular intervals $\Delta \in \real$.
For each point $\bp_i$, we sample the corresponding voxel grid feature $V[\bp_i] \in \real^{d^V}$, where $V[\cdot]$ stands for trilinear interpolation.
The feature $V[\bp_i]$ is then decoded by an MLP as
$
f_\zeta(V[\bp_i], \bru) := (\sigma_i, \bc_i)
$
with parameters $\zeta$ to obtain the density $\sigma_i \in [0, 1]$ and the RGB color $\bc_i \in [0, 1]^3$ of each 3D point.
The MLP $f$ is designed so that the color $\bc$ depends on the ray direction $\bru$ while the density $\sigma$ does not, similar to NeRF \cite{mildenhall2020nerf}.
Finally, EA ray marching renders the $\bru$'s pixel color
$
\bc_{\bru} = \sum_{i=1}^{N_S} w(\bp_i) \bc_i
$ as a weighted combination of the sampled colors.
The weights are defined as
$
w(\bp_i) = T_{i} - T_{i+1}
$
where
$
T_i = e^{-\sum_1^{i-1} \sigma_i \Delta}
$.

%% file: 05_experiments.tex
\section{Experiments}

\begin{figure*}
\centering%
\includegraphics[width=1.02\linewidth]{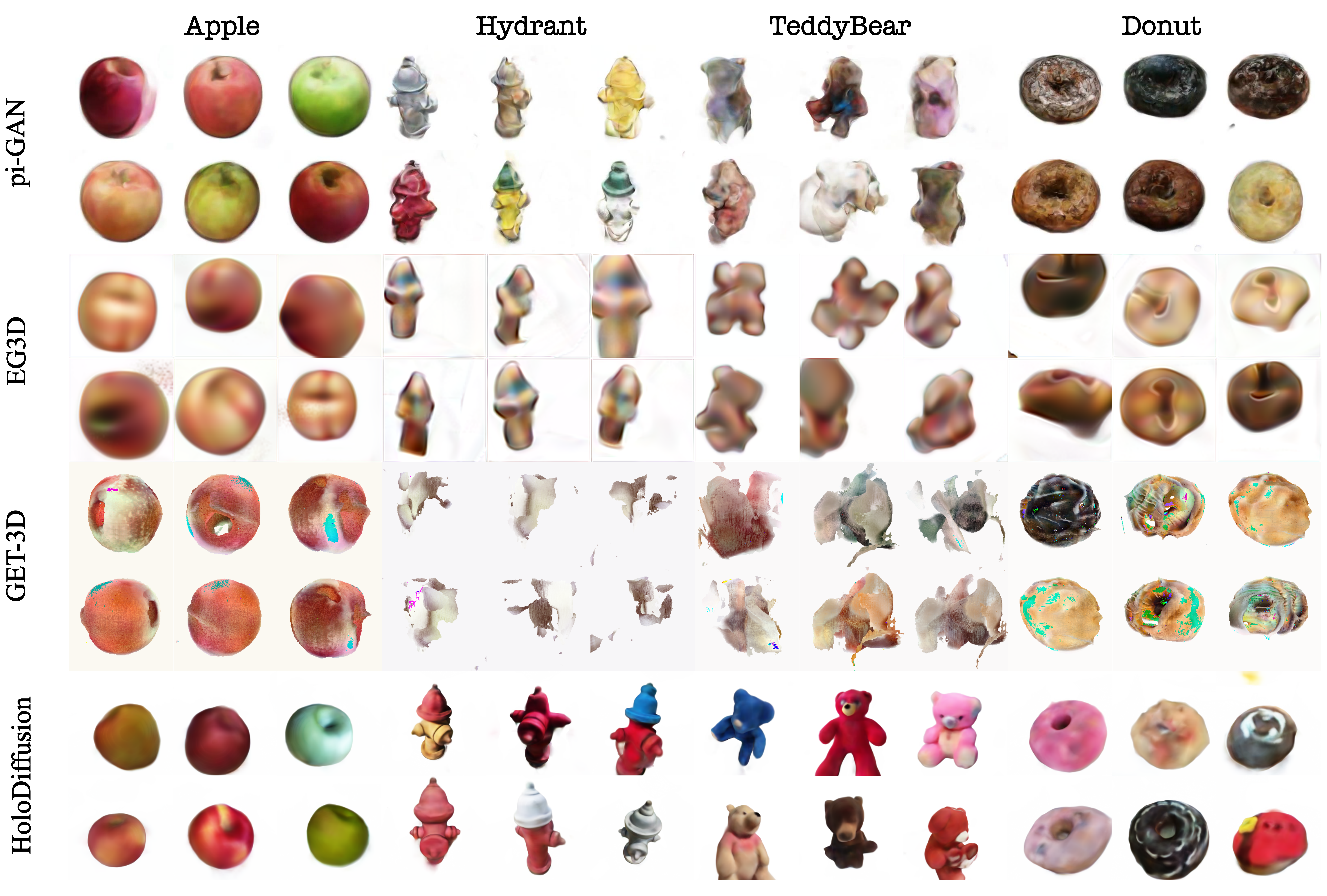}
\caption{
\textbf{Comparisons. }
Samples generated by our \name compared to those by pi-GAN, EG3D, and GET3D. 
\label{fig:mega_qualitative}
\label{fig:qualitative}
}
\end{figure*}

In this section, we evaluate our method.
\new{First} we perform the \textit{quantitative} evaluation and then follow it by visualizing samples for assessing the \textit{quality} of generations.

\paragraph{Datasets and baselines.}

For our experiments, we use CO3Dv2~\cite{reizenstein21common}, which is currently the largest available dataset of fly-around real-life videos of object categories.
The dataset contains videos of different object categories and each video makes a complete circle around the object, showing all sides of it.
Furthermore, camera poses and object foreground masks are provided with the dataset (they were obtained by the authors by running off-the-shelf Structure-from-Motion and instance segmentation software, respectively).

\new{We} consider the four categories \texttt{Apple}, \texttt{Hydrant}, \texttt{TeddyBear} and \texttt{Donut} for our experiments. For each of the categories we train a single model on the 500 ``train'' videos (\new{i.e.} approx. $500 \times 100$ frames in total) with the highest camera cloud quality score, as defined in the CO3Dv2 annotations, in order to ensure clean ground-truth camera pose information. \new{We note} that all trainings were done on $2$-to-$8$ V100 32GB GPUs for 2 weeks. 

\new{We consider} the prior works pi-GAN~\cite{chan2021pi}, EG3D~\cite{chan2022efficient}, and GET3D~\cite{gao2022get3d} as baselines for comparison.
Pi-GAN generates radiance fields represented by MLPs and is trained using an adversarial objective.
Similar to our setting, they only use 2D image supervision for training. 
\new{EG3D~\cite{chan2022efficient}} uses the feature triplane, decoded by an MLP as the underlying representation, while needing both the images and the camera poses as input to the training procedure. \new{GET3D~\cite{gao2022get3d}} is another GAN-based baseline, which also requires the images and camera poses for training. Apart from this, GET3D also requires the fg/bg masks for training; which we supply in form of the masks available in CO3Dv2. Since GET3D applies a Deformable Marching Tetrahedra step in the pipeline, the samples generated by them are in the form of textured meshes.

\paragraph{\new{Quantitative evaluation}.}
\input{___tab_quant}
\begin{figure*}%
\centering%
\includegraphics[width=0.95\linewidth]{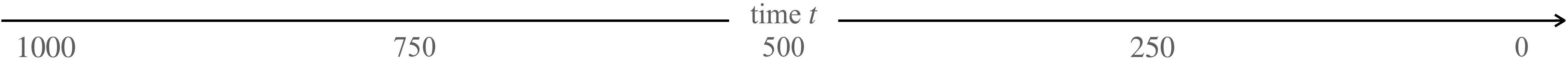}\\
\includegraphics[width=1.02\linewidth]{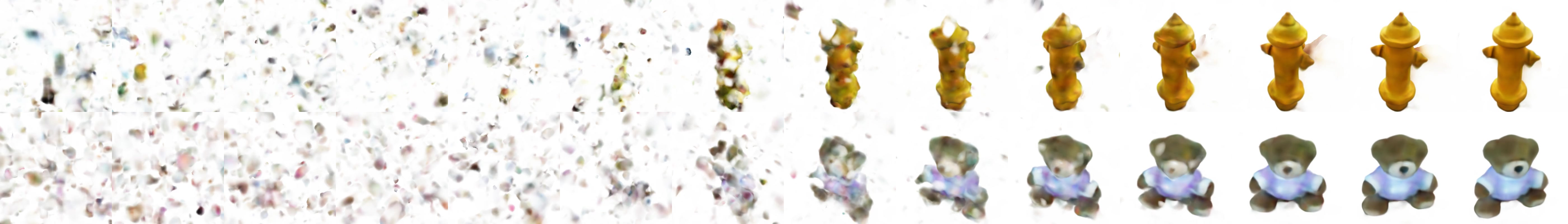}%
\caption{%
\textbf{Sampling across time. }
Rendering of \name's \new{iterative sampling process} for a hydrant and a teddy bear.
The diffusion time decreases from left ($t=T=1000$) to the right ($t=0$).
\label{fig:explosion}
}
\end{figure*}
We report Frechet Inception Distance (FID)~\cite{heusel2017gan-stability}, and Kernel Inception Distance (KID)~\cite{binkowski2018demystifying} for assessing the generative quality of our results. 
As shown in \Cref{tab:base_qunat}, our \name produces better scores than EG3D and GET3D.
Although pi-GAN gets better scores than ours on some categories, we note that the 3D-agnostic training procedure of pi-GAN cannot recover the proper 3D structure of the unaligned shapes of CO3Dv2. Thus, without the 3D-view consistency, the 3D neural fields (MLPs) produced by pi-GAN essentially mimic a 2D image GAN.

\paragraph{\new{Qualitative evaluation}.}
\Cref{fig:qualitative} depicts random samples generated from all the methods under comparison. \name produces the most appealing, consistent and realistic samples among all.
\Cref{fig:qualitative_vp} further analyzes the viewpoint consistency of pi-GAN compared to ours.
It is evident that, although individual views of pi-GAN samples look realistic, their appearance is inconsistent with the change of viewpoint. Please refer to the project webpage
for more examples and videos of the generated samples. %

%% file: ___tab_quant.tex
\begin{table*}[h!]
\centering
\caption{
\textbf{Quantitative evaluation. }
FID and KID on 4 classes of CO3Dv2 comparing our \name with the baselines pi-GAN~\cite{chan2021pi}, \new{EG3D~\cite{chan2022efficient}}, \new{GET3D~\cite{gao2022get3d}}, and the non-bootstrapped version of our \name. \new{The column ``VP'' denotes whether renders of a method are 3D view-consistent or not.}
}
\label{tab:base_qunat}
\resizebox{\linewidth}{!}{
\begin{tabular}{ l  c  rr  rr  rr  rr  rr }
\toprule
 method & VP    & \multicolumn2c{\texttt{\new{Apple}}}        & \multicolumn2c{\texttt{Hydrant}} &
                  \multicolumn2c{\texttt{TeddyBear}}    & \multicolumn2c{\texttt{Donut}}   &
                  \multicolumn2c{Mean} \\
                  \cmidrule(lr){3-4}      \cmidrule(lr){5-6}     \cmidrule(lr){7-8}     \cmidrule(lr){9-10}
                  \cmidrule(lr){11-12}
        &       & \multicolumn1c{\footnotesize{FID $\downarrow$}} & \multicolumn1c{\footnotesize{KID $\downarrow$}} &
                  \multicolumn1c{\footnotesize{FID $\downarrow$}} & \multicolumn1c{\footnotesize{KID $\downarrow$}} &
                  \multicolumn1c{\footnotesize{FID $\downarrow$}} & \multicolumn1c{\footnotesize{KID $\downarrow$}} &
                  \multicolumn1c{\footnotesize{FID $\downarrow$}} & \multicolumn1c{\footnotesize{KID $\downarrow$}} &
                  \multicolumn1c{\footnotesize{FID $\downarrow$}} & \multicolumn1c{\footnotesize{KID $\downarrow$}} \\
\midrule

pi-GAN~\cite{chan2021pi}                        &  \xmark & 49.3  & 0.042 & 92.1  & 0.080 & 125.8  & 0.118 & 99.4  & 0.069
                                                          & 91.7 & 0.077 \\

\new{EG3D}~\cite{chan2022efficient}             &  \cmark & 170.5 & 0.203 & 229.5 & 0.253 & 236.1 & 0.239 & 222.3 & 0.237
                                                          & 214.6 & 0.233 \\

\new{GET3D}~\cite{gao2022get3d}                 &  \cmark & 179.1 & 0.190 & 303.3 & 0.380 & 244.5 & 0.280 & 209.9 & 0.230
                                                          & 234.2 & 0.270 \\
\new{\name (No bootstrap)}                      &  \cmark & 342.9 & 0.400 & 277.9 &	0.305 & 222.1 &	0.217 & 272.1 &	0.199                                                           & 278.7 & 0.280 \\
\name                                           &  \cmark & 94.5     & 0.095 & 100.5 & 0.079 & 109.2  & 0.106 & 115.4 &                                                         0.085 & 104.9  & 0.091 \\
\bottomrule

\end{tabular}}
\end{table*}

%% file: 06_conclusion.tex
\section{Conclusion}
We have presented \name, an unconditional 3D-consistent generative diffusion model that can be trained using only posed-image supervision.
At the core of our method is a learnable rendering module that is trained in conjunction with the diffusion denoiser, which operates directly in the feature space.
Furthermore, we use a pretrained feature encoder to decouple the cubic volumetric memory complexity from the final image rendering resolution.
We demonstrate that the method can be trained on raw posed image sets, even in the few-image setting, striking a good balance between quality and diversity of results. 

At present, our method requires access to camera information at training time. 
One possibility is to jointly train a viewpoint estimator to pose the input images, but the challenge may be to train this module from scratch as the input view distribution is unlikely to be uniform~\cite{Niemeyer2021THREEDV}.
An obvious next challenge would be to test the setup for conditional generation, either based on images (i.e., single view reconstruction task) or using text guidance.
Beyond generation, we would also like to support editing the generated representations, both in terms of shape and appearance, and compose them together towards scene generation.
Finally, we want to explore multi-class training where diffusion models, unlike their GAN counterparts, are known to excel without suffering from mode collapse. 

\section{Societal Impact}
Our method primarily contributes towards the generative modeling of 3D real-captured assets. Thus as is the case with 2D generative models, ours is also prone to misuse of generated synthetic media. In the context of synthetically generated images, our method could potentially be used to make fake 3D view-consistent GIFs or videos. Since we only train our models on the virtually harmless Co3D (Common objects in 3D) dataset, our released models could not be directly used to infer potentially malicious samples.

As diffusion models can be prone to memorizing the training data in limited data settings \cite{somepalli2022diffusion}, our models can also be used to recover the original training samples. Analyzing the severity and the extent to which our models suffer from this, is an interesting future direction for exploration.

%% file: 07_acknowledgements.tex
\section{\new{Acknowledgements}}
Animesh and Niloy were partially funded by the European Union’s Horizon 2020 research and innovation programme under the Marie Skłodowska-Curie grant agreement No.~956585. This research has been partly supported by MetaAI and the UCL AI Centre. Finally, Animesh thanks  \href{https://ajolicoeur.wordpress.com/about/}{Alexia Jolicoeur-Martineau} for the helpful and insightful guidance on diffusion  models.

%% file: suppl_body.tex
\setcounter{figure}{0} \renewcommand{\thefigure}{\Roman{figure}}
\setcounter{table}{0} \renewcommand{\thetable}{\Roman{table}}

\begin{strip}%
 \centering
 \Large
 \textbf{%
{
\name: Training a 3D Diffusion Model using 2D Images
}
\\
 \vspace{0.3cm} 
 \textit{Supplementary material}
 }\\
\vspace{0.5cm}
\end{strip}

\section{Views2Voxel-grid Unprojection Mechanism}
Given a training video $s$ containing frames $I_j$, we generate a grid $\bar V \in \real^{d^V \times S \times S \times S}$
of auxiliary features $\bar V_{:mno} \in [-1, 1]^{d_V}$ by using the following procedure.
We first project the 3D coordinate $\bx^V_{mno}$ of each grid element $(m,n,o)$  to every video frame $I_j$ and sample corresponding 2D image features. The 2D image features $f_{mno}^{j}$ are obtained using a
ResNet32~\cite{he15deep} encoder $E(I_j)$. We use bilinear interpolation for sampling continuous values and use zero-features for projected points that lie outside the Image. Thus, we obtain $N_\text{frames}$ features (corresponding to each frame in the video) for each grid element of the voxel-grid. We accumulate these features using the Accumulator MLP $\mathcal{A}_{acc}$. The accumulator $\mathcal{A}_{acc}$ takes as input $[f_{mno}^{j}; v^j]$, where $[;]$ denotes concatenation and $v^j$ corresponds to the viewing direction corresponding to the camera center of $j^{\text{th}}$ frame, and outputs $[\sigma^j_{mno}; {f'}_{mno}^{j}]$. Finally, we compute the feature at each of the voxel grid centers as a weighted sum of the newly mapped features:
\begin{equation}
    F_{mno} = \sum_{j} \sigma^j_{mno} {f'}_{mno}^{j}.
\end{equation}

\section{Implementation Details}
In this section, we provide more details related to implementing our proposed method.

\subsection{Network Architectures}
Our proposed pipeline (Fig 2. of main paper) contains three neural components: The Encoder, Diffusion UNet and Renderer. The Encoder network is a ResNet32 model \cite{he15deep}.
For the main diffusion network, we use a 3D variant of the UNet used by Dhariwal and Nichol \cite{dhariwal2021diffusion}. The model comprises residual blocks containing downsampling, upsampling, and self-attention blocks (with additive residual connections).

\subsection{Renderer}
\begin{figure}
\centering
\includegraphics[width=0.4\textwidth]{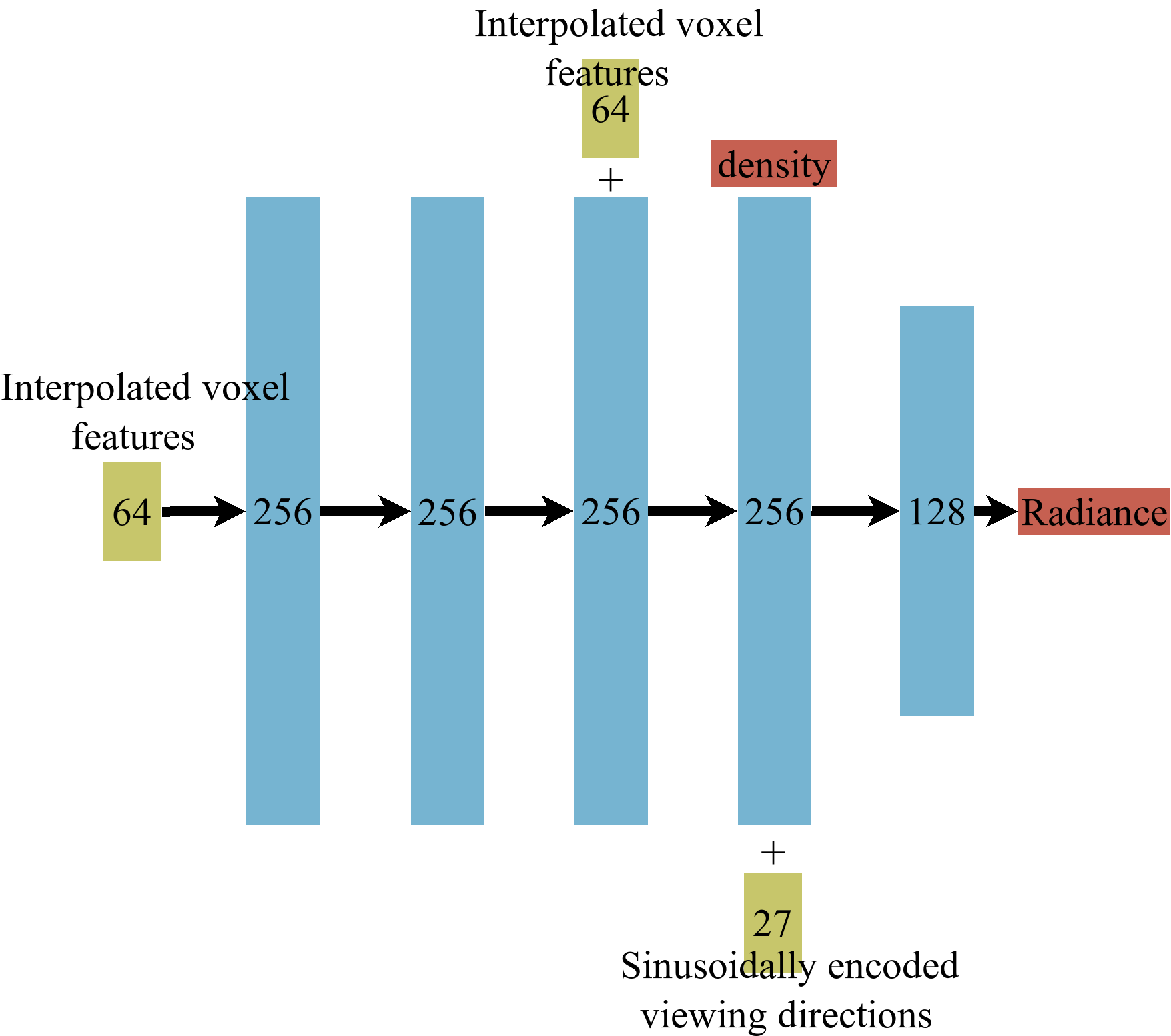}
\caption{Architecture of the \texttt{RenderMLP} used for decoding the features of the generated voxel grids into density and radiance fields.}
\label{fig:render_mlp}
\end{figure}

In order to decode the generated voxel-grid of features into density and radiance fields, we use a NeRF-like \cite{mildenhall2020nerf} MLP (Multi-layer perceptron). The MLP contains 4 layers of 256 hidden units with a skip-connection on the 3rd hidden layer. The skip connection concatenates the input features with the intermediate hidden layer features. Similar to NeRF, and for the reasons described in Zhang et al. \cite{zhang2020nerf++}, we also input the view-directions at a latter layer in the MLP. The input features are not encoded, but we apply sinusoidal encodings \cite{mildenhall2020nerf, vaswani17attention} to the input viewing directions with max frequency level $L=4$. The activation functions used are: \texttt{LeakyReLU} for the hidden layers, \texttt{Softplus} for the density output head, and the \texttt{Sigmoid} for the radiance output head. All trainable weights are initialized using the Xavier uniform initialization \cite{glorot2010understanding}. Figure \ref{fig:render_mlp} shows the detailed architecture of the \texttt{RenderMLP}. 

\subsection{Training Details}
We train the full \name pipeline for $1000$ epochs over the dataset containing the object-centric videos. During training, we randomly sample 11 source views for unprojecting into the initial voxel-grid, and 1 target (reserved) novel view for computing loss. The latter enforces 3D structure in the generated samples. We use $L2$ distance between the rendered views and the G.T. views as the photometric-consistency loss. In terms of hardware, we train all our models on 4-8 32GB-V100 GPUs, with a \texttt{batch-size} equal to the number of GPUs in use, i.e., each GPU processes one voxel-grid during training.
We use \textit{Adam} \cite{kingma15adam:} optimizer with a learning rate ($\alpha$) of $0.00005$ and default values of $\beta_1$, $\beta_2$, and $\epsilon$ for all the trainable networks during training.

\subsection{Diffusion Details}
We use the DDPM \cite{ho2020denoising} diffusion-formulation for our bootstrap-latent-diffusion module as described in section 4.2 of the main paper. We use the default $t = 1000$ time-steps and the default $\beta_t$ schedule in our experiments: wherein we set $\beta_0 = 0.0001; \beta_{999} = 0.02$. Rest of the $\beta_t$ values are obtained by linearly interpolating between the $\beta_0$ and $\beta_{999}$. Finally, to improve the input conditioning of our diffusion module, we apply $tanh$ to the voxel features to constrain their values in the range of [-1, 1], as proposed in Karras et al. \cite{karras2022elucidating}. This allows us to apply [-1, 1] clipping during sampling.